# Conditional Chow-Liu Tree Structures for Modeling Discrete-Valued Vector Time Series

**Sergey Kirshner, Padhraic Smyth,**
School of Information and Computer Science
University of California, Irvine
Irvine, CA 92697-3425
{skirshne,smyth}@ics.uci.edu

**Andrew W. Robertson**
International Research Institute
for Climate Prediction (IRI),
The Earth Institute at Columbia University,
Palisades, NY 10964
awr@iri.columbia.edu

## Abstract

We consider the problem of modeling discrete-valued vector time series data using extensions of Chow-Liu tree models to capture both dependencies across time and dependencies across variables. Conditional Chow-Liu tree models are introduced, as an extension to standard Chow-Liu trees, for modeling conditional rather than joint densities. We describe learning algorithms for such models and show how they can be used to learn parsimonious representations for the output distributions in hidden Markov models. These models are applied to the important problem of simulating and forecasting daily precipitation occurrence for networks of rain stations. To demonstrate the effectiveness of the models, we compare their performance versus a number of alternatives using historical precipitation data from Southwestern Australia and the Western United States. We illustrate how the structure and parameters of the models can be used to provide an improved meteorological interpretation of such data.

## 1 Introduction

In this paper we consider the problem of modeling discrete-time, discrete-valued, multivariate time-series. For example, consider $M$ time-series where each time-series can take $B$ values. The motivating application in this paper is modeling of daily binary rainfall data ($B = 2$) for regional spatial networks of $M$ stations (where typically $M$ can vary from 10 to 100). Modeling and prediction of rainfall is an important problem in the atmospheric sciences. A common application, for example, is simulating realistic daily rainfall patterns for a 90-day season, to be used as input for detailed crop-modeling simulations. A number of statistical methods have been developed for modeling daily rainfall time-series at single stations—first-order Markov models and various extensions (also known as "weather generators") have proven quite effective for single-station rainfall modeling in many geographic regions. However, there has been less success in developing models for multiple stations that can generate simulations with realistic spatial and temporal correlations in rainfall patterns (Wilks and Wilby 1999).

Direct modeling of the dependence of the $M$ daily observations at time $t$ on the $M$ observations at time $t-1$ requires an exponential number of parameters in $M$. This is clearly impractical for most values of $M$ of interest. In this paper we look at the use of hidden Markov models (HMMs) to avoid this problem—an HMM uses a $K$-valued hidden first-order Markov chain to model time-dependence, with the $M$ outputs at time $t$ being conditionally independent of everything else given the current state value at time $t$. The hidden state variable in an HMM serves to capture temporal dependence in a low-dimensional manner, i.e., with $O(K^2)$ parameters instead of being exponential in $M$. From a scientific viewpoint, an attractive feature of the HMM is that the hidden states can be interpreted as underlying "weather states" (Hughes et al. 1999, Robertson et al. to appear).

Modeling the instantaneous multivariate dependence of the $M$ observations on the state at time $t$ would require $B^M$ parameters per state if the full joint distribution were modeled. This in turn would defeat the purpose of using the HMM to reduce the number of parameters. Thus, approximations such as assuming conditional independence (CI) of the $M$ observations are often used in practice (e.g., Hughes et al. 1999), requiring $O(KMB)$ parameters.

While the HMM-CI approach is a useful starting point it suffers from two well-known disadvantages for an application such as rainfall modeling: (1) the assumed conditional independence of the $M$ outputs on each other at time $t$ can lead to inadequate characterization of the dependence between the $M$ time-series (e.g., unrealistic spatial rainfall patterns on a given day), (2) the assumed conditional independence of the $M$ outputs at time $t$ from from the $M$ outputs at time $t-1$ can lead to inadequate temporal dependence in the $M$ time-series (e.g., unrealistic occurrences of wet days during dry spells).

In this paper we investigate Chow-Liu tree structures in the context of providing improved, yet tractable, models to address these problems in capturing output dependencies for HMMs. We show how Chow-Liu trees can be used to directly capture dependency among the $M$ outputs in multivariate HMMs. We also introduce an extension called conditional Chow-Liu trees to provide a class of dependency models that are well-suited for modeling multivariate time-series data. We illustrate the application of the proposed methods to two large-scale precipitation data sets.

The paper is structured as follows. Section 2 formally describes existing models and our extensions. Section 3 describes how to perform inference and to learn both the structure and parameters for the models. Section 4 describes an application and analyzes the performance of the models. Finally, Section 5 summarizes our contributions and outlines possible future directions.

## 2 Model Description

We begin this section by briefly reviewing Chow-Liu trees for multivariate data before introducing the conditional Chow-Liu tree model. We then focus on vector time-series data and show how the conditional Chow-Liu tree model and hidden Markov models can be combined.

### 2.1 Chow-Liu Trees

Chow and Liu (1968) proposed a method for approximating the joint distribution of a set of discrete variables using products of distributions involving no more than pairs of variables. If $P(\mathbf{x})$ is an $M$-variate distribution on discrete variables $V = \left(x^1, \ldots, x^M\right)$, the Chow-Liu method constructs a distribution $T(\mathbf{x})$ for which the corresponding Bayesian and Markov network is tree-structured. If $G_T = (V, E_T)$ is the Markov network associated with $T$, then

$$T(\mathbf{x}) = \frac{\prod_{(u,v)\in E_T} T(x^u, x^v)}{\prod_{v\in V} T(x^v)^{degree(v)}} = \prod_{(u,v)\in E_T} \frac{T(x^u, x^v)}{T(x^v)\,T(x^u)} \prod_{v\in V} T(x^v). \quad (1)$$

The Kullback-Leibler divergence $KL(P,T)$ between

Algorithm ChowLiu($P$)
**Inputs:** Distribution $P$ over domain $V$; procedure MWST( weights ) that outputs a maximum weight spanning tree over $V$

1. Compute marginal distributions $P(x^u, x^v)$ and $P(x^u)$ $\forall u, v \in V$
2. Compute mutual information values $I(x^u, x^v)$ $\forall u, v \in V$
3. $E_T = \text{MWST}(\{I(x^u, x^v)\})$
4. Set $T(x^u, x^v) = P(x^u, x^v)$ $\forall (u,v) \in E_T$

**Output:** $T$

Figure 1: Chow-Liu algorithm (very similar to Meilă and Jordan 2000)

distributions $P$ and $T$ is defined as

$$KL(P,T) = \sum_{\mathbf{x}} P(\mathbf{x}) \log \frac{P(\mathbf{x})}{T(\mathbf{x})}.$$

Chow and Liu showed that in order to minimize $KL(P,T)$ the edges for the tree $E_T$ have to be selected to maximize the total mutual information of the edges $\sum_{(u,v)\in E_T} I(x^u, x^v)$ where mutual information between variables $x^u$ and $x^v$ is defined as

$$I(x^u, x^v) = \sum_{x^u}\sum_{x^v} P(x^u, x^v) \log \frac{P(x^u, x^v)}{P(x^u)\,P(x^v)}. \quad (2)$$

This can be accomplished by calculating mutual information $I(x^u, x^v)$ for all possible pairs of variables in $V$, and then solving the maximum spanning tree problem, with pairwise mutual information from Equation 2 as edge weights (e.g., Cormen et al. 1990). Once the edges are selected, the probability distribution $T$ on the pairs of vertices connected by edges is defined to be the same as $P$:

$$\forall (x^u, x^v) \in E_T \quad T(x^u, x^v) = P(x^u, x^v),$$

and the resulting distribution $T$ minimizes $KL(P,T)$. Figure 1 outlines the algorithm for finding $T$.

If each of the variables in $V$ takes on $B$ values, finding the tree $T$ has complexity $O\left(M^2B^2\right)$ for the mutual information calculations and $O\left(M^2\right)$ for finding the minimum spanning tree, totaling $O\left(M^2B^2\right)$. (Meilă (1999) proposed a faster version of the Chow-Liu algorithm for sparse high-dimensional data.) In practice, $P$ is often an empirical distribution on the data, so that the calculation of pairwise counts of variables (used in calculating mutual information) has complexity $O\left(TM^2B^2\right)$ where $T$ is the number of vectors in the data.

The advantages of Chow-Liu trees include (a) the existence of a simple algorithm for finding the optimal tree, (b) the parsimonious nature of the model (the number of parameters is linear in the dimensionality of the space), and (c) the tree structure $T$ can have a simple intuitive interpretation. While there are other algorithms that retain the idea of a tree-structured distribution, while allowing for more complex dependencies (e.g., thin junction trees, Bach and Jordan 2002), these algorithms have higher time complexity than the original Chow-Liu algorithm and do not guarantee optimality within the model class for the structure that is learned. Thus, in the results in this paper we focus on Chow-Liu trees under the assumption that they are a generally useful modeling technique.

### 2.2 Conditional Chow-Liu Forests

It is common in practice (e.g., in time-series and in regression modeling) that the data to be modeled can be viewed as consisting of two sets of variables, where we wish to model the conditional distribution $P(\mathbf{x}|\mathbf{y})$ of one set $\mathbf{x}$ on the other set $\mathbf{y}$. We propose an extension of the Chow-Liu method to model such conditional distributions. As with Chow-Liu trees, we want the corresponding probability distribution to be factored into a product of distributions involving no more than two variables. Pairs of variables are represented as an edge in a corresponding graph with nodes corresponding to variables in $V = V_x \cup V_y$. However, since all of the variables in $V_y$ are observed, we are not interested in modeling $P(\mathbf{y})$, and do not wish to restrict $P(\mathbf{y})$ by making independence assumptions about the variables in $V_y$. The structure for an approximation distribution $T$ will be constructed by adding edges such as not to introduce paths involving multiple variables from $V_y$.

Let $G_F = (V, E_F)$ be a forest, a collection of disjoint trees, containing edges $E_x$ between pairs of variables in $V_x$ and edges $E_y$ connecting variables from $V_x$ and $V_y$, $E_F = E_x \cup E_y$. If the probability distribution $T(\mathbf{x}|\mathbf{y})$ has $G_F$ for a Markov network, then similar to Equation 1:

$$\begin{aligned} T(\mathbf{x}|\mathbf{y}) &= \prod_{(u,v)\in E_x} \frac{T(x^u, x^v)}{T(x^u)\,T(x^v)} \prod_{v\in V_x} T(x^v) \\ &\quad \times \prod_{(u,v)\in E_y} \frac{T(y^u, x^v)}{T(y^u)\,T(x^v)}. \end{aligned}$$

We will again use KL-divergence between conditional distributions $P$ and $T$ as an objective function:

$$KL(P,T) = \sum_{\mathbf{y}} P(\mathbf{y}) \sum_{\mathbf{x}} P(\mathbf{x}|\mathbf{y}) \log \frac{P(\mathbf{x}|\mathbf{y})}{T(\mathbf{x}|\mathbf{y})}.$$

It can be shown that the optimal probability distribution $T$ with corresponding Markov network $G_F$ is

$$\forall (u,v) \in E_x,\ T(x^u, x^v) = P(x^u, x^v)$$

and

$$\forall (u,v) \in E_y,\ T(y^u, x^v) = P(y^u, x^v).$$

As with the unconditional distribution, we wish to find pairs of variables to minimize

$$\begin{aligned} KL(P,T) &= \sum_{v\in V_x} H[x^v] - H[\mathbf{x}|\mathbf{y}] \\ &\quad - \sum_{(u,v)\in E_x} I(x^u, x^v) - \sum_{(u,v)\in E_y} I(y^u, x^v) \end{aligned}$$

where $H[x^v]$ denotes the entropy of $P(x^v)$, and $H[\mathbf{x}|\mathbf{y}]$ denotes the conditional entropy of $P(\mathbf{x}|\mathbf{y})$. Both $H[\mathbf{x}]$ and $H[\mathbf{x}|\mathbf{y}]$ are independent of $E_F$, so as in the unconditional case, we need to solve a maximum spanning tree problem on the graph with nodes $V_y \cup V_x$ while not allowing paths between vertices in $V_y$ (alternatively, assuming all nodes in $V_y$ are connected).

The algorithm for learning the conditional Chow-Liu (CCL) distribution is shown in Figure 2. Due to the restrictions on the edges, the CCL networks can contain

Algorithm CondChowLiu($P$)
**Inputs:** Distribution $P$ over domain $V_x \cup V_y$; procedure MWST( $V$, weights ) that outputs a maximum weight spanning tree over $V$

1. (a) Compute marginal distributions $P(x^u, x^v)$ and $P(x^u)$ $\forall u, v \in V_x$
   (b) Compute marginal distributions $P(y^u)$ and $P(y^u, x^v)$ $\forall u \in V_y, v \in V_x$
2. (a) Compute mutual information values $I(x^u, x^v)$ $\forall u, v \in V_x$
   (b) Compute mutual information values $I(y^u, x^v)$ $\forall u \in V_y, v \in V_x$
   (c) Find $u(v) = \arg\max_{u\in V_y} I(y^u, x^v)$ $\forall v \in V_x$
   (d) Let $V' = V_x \cup \{v'\}$, and set $I\left(x^{v'}, x^v\right) = I\left(y^{u(v)}, x^v\right)$ $\forall v \in V_x$
3. (a) $E_{T'} = \text{MWST}(V', \mathbf{I})$
   (b) $E_x = \{(u,v) \,|u, v \in V_x, (u,v) \in E_{T'}\}$
   (c) $E_y = \{(u(v), v) \,|v \in V_x, (v, v') \in E_{T'}\}$.
4. (a) Set $T(x^u, x^v) = P(x^u, x^v)$ $\forall (u,v) \in E_x$
   (b) Set $T(y^u, x^v) = P(y^u, x^v)$ $\forall (u,v) \in E_y$

**Output:** $T$

Figure 2: Conditional Chow-Liu algorithm

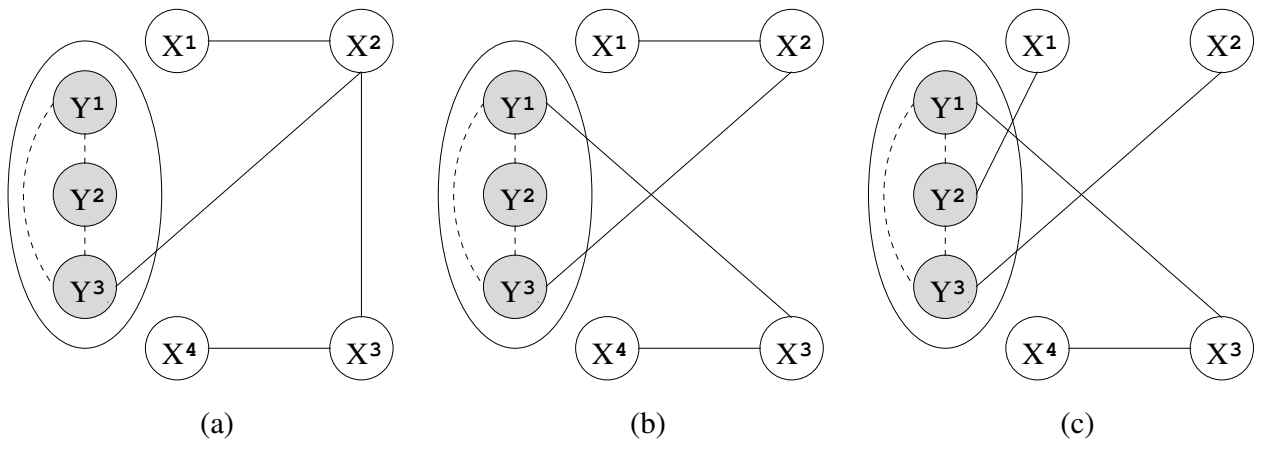

Figure 3: Conditional CL forest for a hypothetical distribution with (a) 1 component (b) 2 components (c) 3 components.

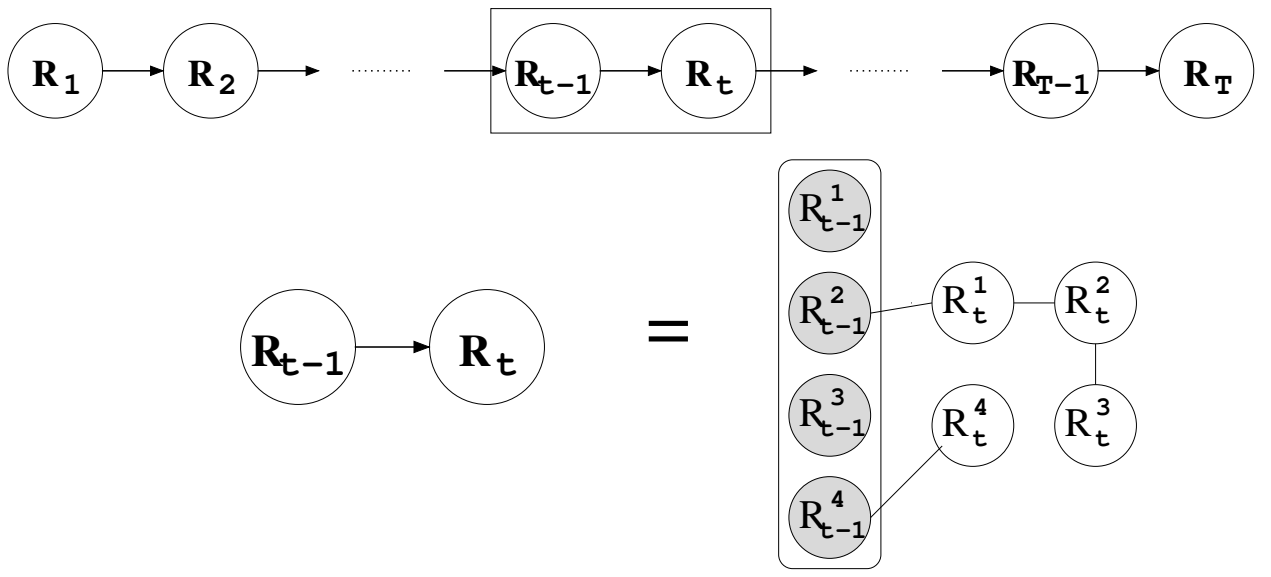

Figure 4: Graphical model for a hypothetical CCLF

disconnected tree components (referred to as forests). These CCL forests can consist of 1 to $\min\{|V_y|, |V_x|\}$ components. (See Figure 3 for an illustration.)

### 2.2.1 Chain CL Forests

We now return to our original goal of modeling time-dependent data. Let $\mathbf{R}_t = \left(R_t^1, \ldots, R_t^M\right)$ be a multivariate ($M$-variate) random vector of data with each component taking on values $\{0, \ldots, B-1\}$. By $\mathbf{R}_{1:T}$ we will denote observation sequences of length $T$.

A simple model for such data can be constructed using conditional Chow-Liu forests. For this chain Chow-Liu forest model (CCLF), the data for a time point $t$ is modeled as a conditional Chow-Liu forest given data at point $t-1$ (Figure 4):

$$P\left(\mathbf{R}_{1:T}\right) = \prod_{t=1}^{T} T\left(\mathbf{R}_t | \mathbf{R}_{t-1}\right)$$

where

$$\begin{aligned} &T\left(\mathbf{R}_t = \mathbf{r} | \mathbf{R}_{t-1} = \mathbf{r}'\right) = \\ &= \prod_{(u,v)\in E_V} \frac{T\left(R_t^u = r^u, R_t^v = r^v\right)}{T\left(R_t^v = r^v\right) T\left(R_t^u = r^u\right)} \prod_{v \in R_t} T\left(R_t^v = r^v\right) \\ &\times \prod_{(u,v)\in E_{Ii}} \frac{T\left(R_t^v = r^v | R_{t-1}^u = r'^u\right)}{T\left(R_t^v = r^v\right)}. \end{aligned}$$

Note that learning the structure and parameters of CCLF requires one pass through the data to collect the counts and calculate joint probabilities of the pairs of variables, and only one run of the CondChowLiu tree algorithm.

## 2.3 Hidden Markov Models

An alternative approach to modeling $\mathbf{R}_{1:T}$ is to use a hidden-state model to capture temporal dependence. Let $S_t$ be the hidden state for observation $t$, taking on one of $K$ values from 1 to $K$, where $S_{1:T}$ denotes sequences of length $T$ of hidden states.

A first-order HMM makes two conditional independence assumptions. The first assumption is that the hidden state process, $S_{1:T}$, is first-order Markov:

$$P\left(S_t | S_{1:t-1}\right) = P\left(S_t | S_{t-1}\right) \tag{3}$$

and that this first-order Markov process is homogeneous in time, i.e., the $K \times K$ transition probability matrix for Equation 3 does not change with time.

The second assumption is that each vector $\mathbf{R}_t$ at time $t$ is independent of all other observed and unobserved states up to time $t$, conditional on the hidden state $S_t$ at time $t$ , i.e.,

$$P\left(\mathbf{R}_t | S_{1:t}, \mathbf{R}_{1:t-1}\right) = P\left(\mathbf{R}_t | S_t\right). \tag{4}$$

Specifying a full joint distribution $P\left(\mathbf{R}_t | S_t\right)$ would require $O(B^M)$ joint probabilities per state, which is clearly impractical even for moderate values of $M$. In practice, to avoid this problem, simpler models are often used, such as assuming that each vector component $R_t^j$ is conditionally independent (CI) of the other components, given the state $S_t$, i.e., $P\left(\mathbf{R}_t | S_t\right) = P(R_t^1, \ldots, R_t^M | S_t) = \prod_{j=1}^{M} P(R_t^j | S_t)$. We will use this HMM-CI as our baseline model in the experimental results section later in the paper—in what follows below we explore models that can capture more dependence structure by using CL-trees.

## 2.4 Chow-Liu Structures and HMMs

We can use HMMs with Chow-Liu trees or conditional Chow-Liu forests to model the output variable given the hidden state. HMMs can model temporal structure of the data while the Chow-Liu models can capture "instantaneous" dependencies between multivariate outputs as well as additional dependence between vector components at consecutive observations over time that the state variable does not capture.

By combining HMMs with the Chow-Liu tree model and with the conditional Chow-Liu forest model we obtain HMM-CL and HMM-CCL models, respectively. The set of parameters $\mathbf{\Theta}$ for these models with $K$ hidden states and $B$-valued $M$-variate vector sets consists of a $K \times K$ transition matrix $\mathbf{\Gamma}$, a $K \times 1$ vector $\mathbf{\Pi}$ of probabilities for the first hidden state in a sequence, and Chow-Liu trees or conditional forests for each hidden state $\mathbf{T} = \{T_1, \ldots, T_K\}$. Examples of graphical model structures for both the HMM-CL and HMM-CCL are shown in Figure 5. The likelihood of $\mathbf{\Theta}$ can then be computed as

$$\begin{aligned} L(\mathbf{\Theta}) &= P(\mathbf{R}_{1:T}|\mathbf{\Theta}) = \sum_{S_{1:T}} P(S_{1:T}, \mathbf{R}_{1:T}|\mathbf{\Theta}) \\ &= \sum_{S_{1:T}} P(S_1|\mathbf{\Theta}) \prod_{t=2}^{T} P(S_t|S_{t-1}, \mathbf{\Theta}) \\ &\quad \times \prod_{t=1}^{T} P(\mathbf{R}_t|S_t, \mathbf{R}_{t-1}, \mathbf{\Theta}) \\ &= \sum_{i_1=1}^{K} \pi_{i_1} T_{i_1}(\mathbf{R}_1) \sum_{t=2}^{T} \sum_{i_t=1}^{K} \gamma_{i_{t-1} i_t} T_{i_t}(\mathbf{R}_t|\mathbf{R}_{t-1}) \end{aligned}$$

with $P(\mathbf{R}_t|S_t, \mathbf{R}_{t-1}, \mathbf{\Theta}) = P(\mathbf{R}_t|S_t, \mathbf{\Theta})$ and $T_i(\mathbf{R}_t|\mathbf{R}_{t-1}) = T_i(\mathbf{R}_t)$ for the HMM-CL.

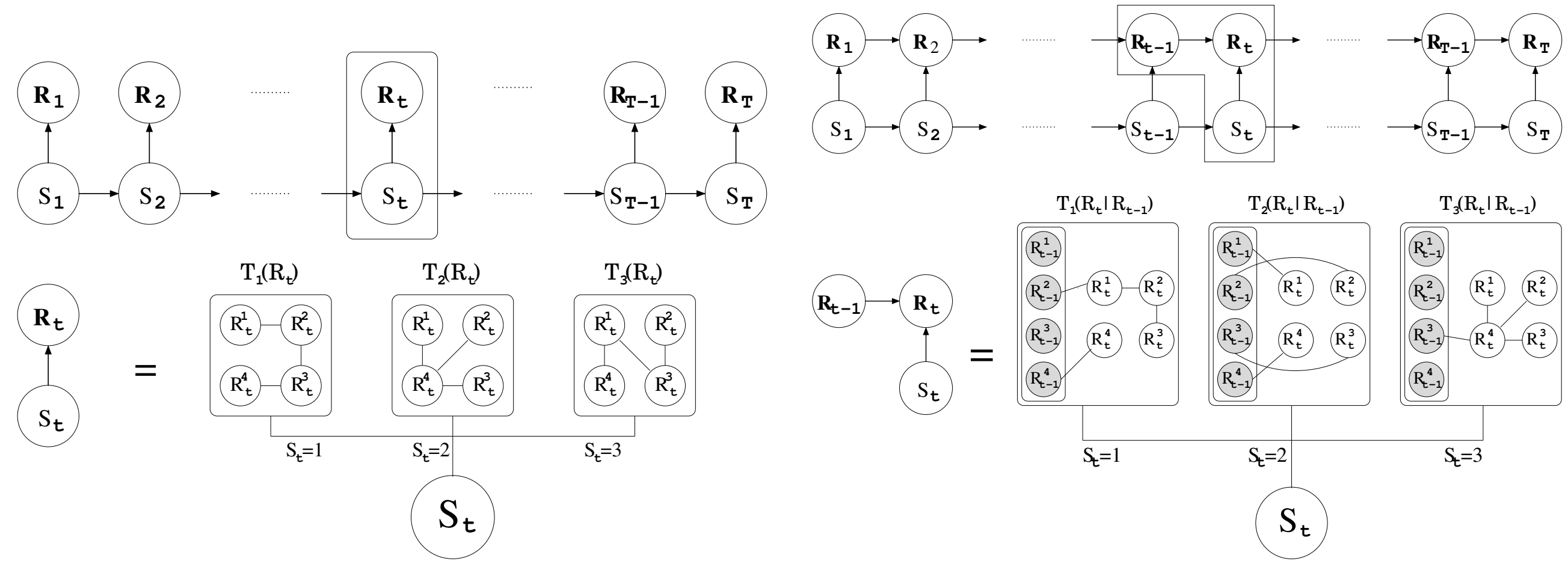

Figure 5: Graphical model interpretation of a hypothetical HMM-CL (left) and HMM-CCL (right)

For the value for the hidden state $S_{t-1} = i$, the probability distribution $P(\mathbf{R}_t|\mathbf{\Theta})$ is just a mixture of Chow-Liu trees (Meilă and Jordan 2000) with mixture coefficients $(\gamma_{i1}, \ldots, \gamma_{iK})$ equal to the $i$-th row of the transition matrix $\Gamma$.

As a side note, since the output part of the HMM-CCL contains dependencies on observations at the previous time-step, the model can be viewed as a form of autoregressive HMM (Rabiner 1989).

## 3 Inference and Learning of HMM-based Models

In this section we discuss both (a) learning the structure and the parameters of the HMM-CL and HMM-CCL models discussed above, and (b) inferring probability distributions of the hidden states for given a set of observations and a model structure and its parameters. We outline how these operations can be performed for the HMM-CL and HMM-CCL. For full details, see Kirshner et al. (2004).

### 3.1 Inference of the Hidden State Distribution

The probability of the hidden variables $S_{1:T}$ given complete observations $\mathbf{R}_{1:T}$ can be computed as

$$P(S_{1:T}|\mathbf{R}_{1:T}) = \frac{P(S_{1:T}, \mathbf{R}_{1:T})}{\sum_{S_{1:T}} P(S_{1:T}, \mathbf{R}_{1:T})}.$$

The likelihood (denominator) cannot be calculated directly since the sum is exponential in $T$. However, the well-known recursive Forward-Backward procedure can be used to collect the necessary information in $O(TK^2M)$ without exponential complexity (e.g., Rabiner 1989). Since the Forward-Backward algorithm for HMM-CLs and HMM-CCLs is not very different from standard HMMs, we will omit the details.

### 3.2 Learning

Learning in HMMs is typically performed using the Baum-Welch algorithm (Baum et al. 1970), a variant of the Expectation-Maximization (EM) algorithm (Dempster et al. 1977). Each iteration of EM consists of two steps. First (E-step), the estimation of the posterior distribution of latent variables is accomplished by the Forward-Backward routine. Second (M-step), the parameters of the models are updated to maximize the expected log-likelihood of the model given the distribution from the M-step. The structures of the trees are also updated in the M-step.

The parameters $\Pi$ and $\Gamma$ are calculated in the same manner as for regular HMMs. Updates for $T_1, \ldots, T_K$ are computed in a manner similar to that for mixtures of trees (Meilă and Jordan 2000). Suppose $\mathbf{R}_{1:T} = \mathbf{r}_{1:T}$. Let $T'_i$ denote the Chow-Liu tree for $S_t = i$ under the updated model. It can be shown (Kirshner et al. 2004) that to improve the log-likelihood one needs to maximize

$$\sum_{i=1}^{K} \left( \sum_{\tau=1}^{T} P\left(S_\tau = i | \mathbf{R}_{1:T} = \mathbf{r}_{1:T}\right) \right) \sum_{t=1}^{T} P_i\left(\mathbf{r}_t\right) \log T'_i\left(\mathbf{r}_t\right)$$

where $P_i\left(\mathbf{r}_t\right) = \frac{P(S_t=i|\mathbf{R}_{1:T}=\mathbf{r}_{1:T})}{\sum_{\tau=1}^{T} P(S_\tau=i|\mathbf{R}_{1:T}=\mathbf{r}_{1:T})}$. This can be accomplished by separately learning Chow-Liu structures for the distributions $P_i$, the normalized posterior distributions of the hidden states calculated in the E-step. The time complexity for each iteration is then $O\left(TK^2M\right)$ for the E-step and $O\left(TK^2 + KTM^2B^2\right)$ for the M-step.

## 4 Experimental Results

To demonstrate the application of the HMM-CL and HMM-CCL models, we consider the problem of modeling precipitation occurrences for a network of rain stations. The data we examine here consists of binary measurements (indicating precipitation or not) recorded each day over a number of years for each of a set of rain stations in a local region. Figure 6 shows a network of such stations in Southwestern Australia.

The goal is to build models that broadly speaking capture both the temporal and spatial properties of the precipitation data. These models can then be used to simulate realistic rainfall patterns over seasons (e.g., 90-day sequences), as a basis for making seasonal forecasts (Robertson et al. to appear), and to fill in missing rain station reports in the historical record.

Markov chains provide a well-known benchmark for modeling precipitation time-series at individual stations (e.g., Wilks and Wilby 1999). However, it is non-trivial to couple multiple chains together so that they exhibit realistic spatial correlation in simulated rainfall patterns. We also compare against the simplest HMM with a conditional independence (CI) assumption for the rain stations given the state. This model captures the marginal dependence of the stations to a certain degree since (for example) in a "wet state" the probability for all stations to be wet is higher, and so forth. However, the CI assumption clearly does not fully capture the spatial dependence, motivating the use of models such as HMM-CL and HMM-CCL.

In the experiments below we use data from both

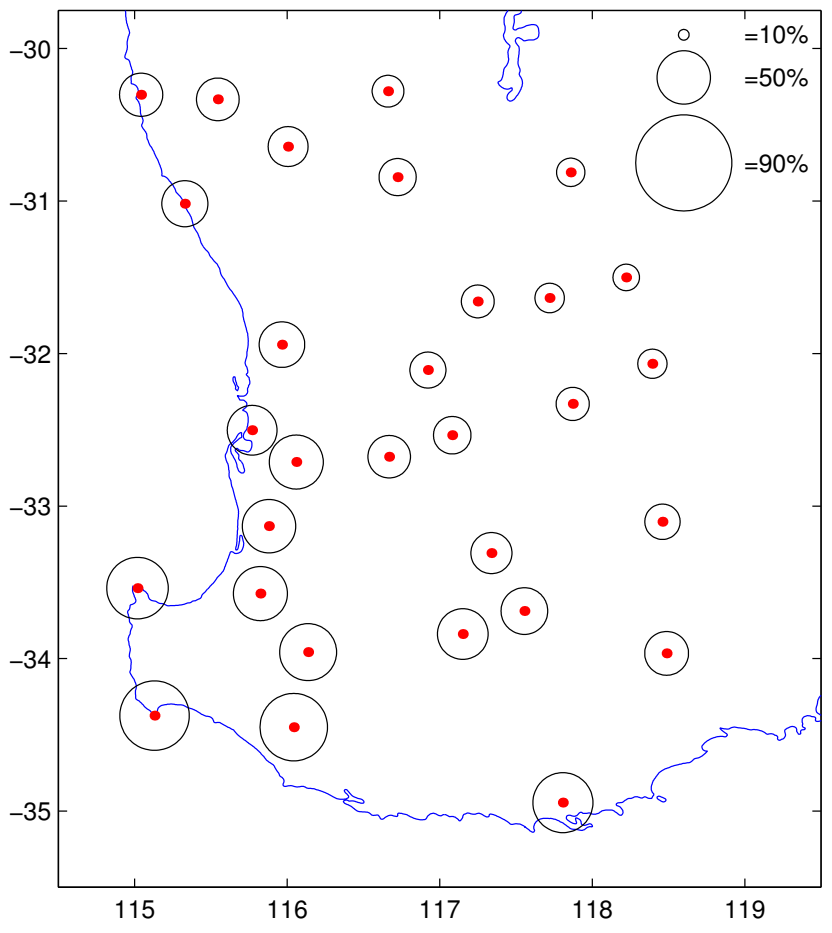

Figure 6: Stations in the Southwestern Australia region. Circle radii indicate marginal probabilities of rainfall ($> 0.3mm$) at each location.

Southwestern Australia (30 stations, 15 184-day winter seasons beginning May 1) and the Western United States (8 stations, 39 90-day seasons beginning December 1). In fitting HMMs to this type of precipitation data the resulting "weather states" are often of direct scientific interest from a meteorological viewpoint. Thus, in evaluating these models, models that can explain the data with fewer states are generally preferable.

We use leave-one-out cross-validation to evaluate the fit of the models to the data. For evaluation we use two different criteria. We compute the log-likelihood for seasons not in the training data, normalized by the number of binary events in the left-out sets (referred to here as out-of-sample scaled log-likelihood). We also compute the average classification error in predicting observed randomly-selected station readings that are deliberately removed from the training data and then predicted by the model. The models considered are the independent Markov chains model (or "weather generator" model), the chain Chow-Liu forest model, the HMM with conditional independence (HMM-CI), the HMM with Chow-Liu tree emissions (HMM-CL), and the HMM with conditional Chow-Liu tree emissions (HMM-CCL). For HMMs, $K$ is chosen corresponding to the largest out-of-sample scaled log-likelihood for each model—the smallest such $K$ is then used across different HMM types for comparison.

The scatter plots in Figures 7 and 8 show the out-of-sample scaled log-likelihoods and classification errors for the models on the left-out sets. The y-axis is the performance of the HMM-CCL model, and the x-axis represents the performance of the other models (shown with different symbols). Higher implies better performance for log-likelihood (on the left) and worse for error (on the right). The HMM-CL and HMM-CCL models are systematically better than the CCLF and HMM-CI models, for both score functions, and for both data sets. The HMM-CCL model does relatively better than the HMM-CL model on the U. S. data. This is explained by the fact that the Australian stations are much closer spatially than the U.S. stations, so that for the U.S. the temporal connections that the HMM-CCL adds are more useful than the spatial connections that the HMM-CL model is limited to.

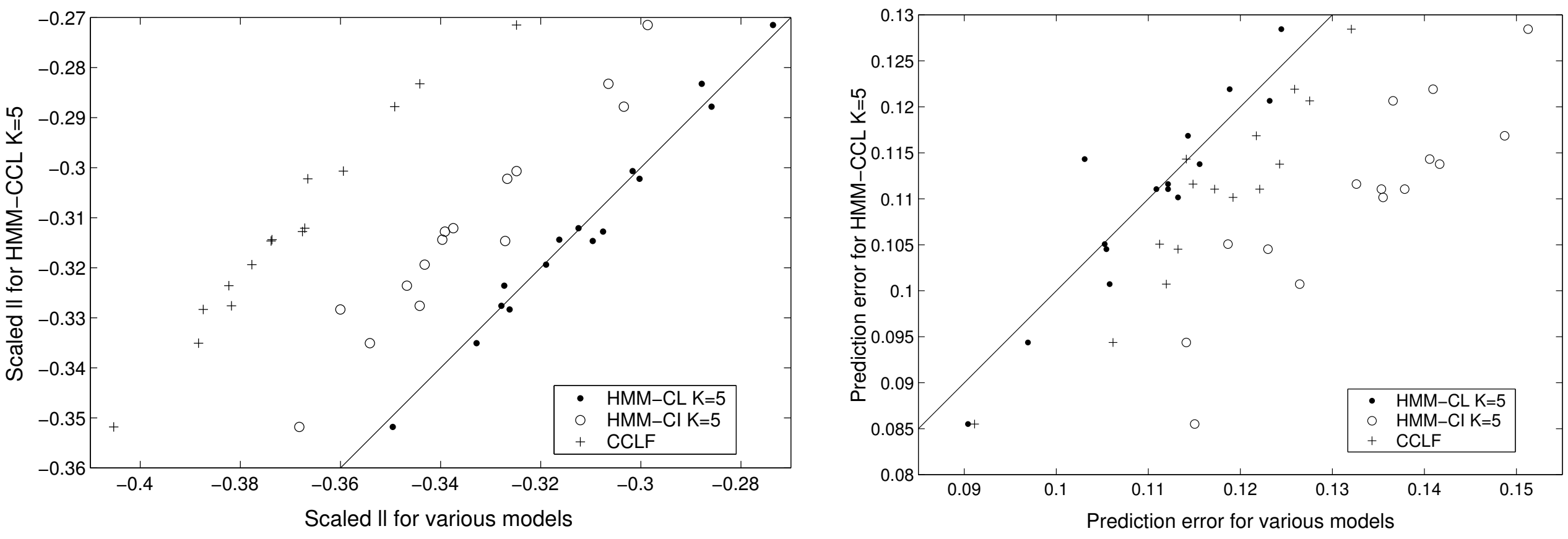

Figure 7: Southwestern Australia data: scatterplots of out-of-sample scaled log-likelihoods (left) and average prediction error (right) obtained by leave-one-winter-out cross-validation. The line corresponds to $y = x$. The independent chains model is not shown since it is beyond the range of the plot (average $ll = -0.6034$, average error $= 0.291$).

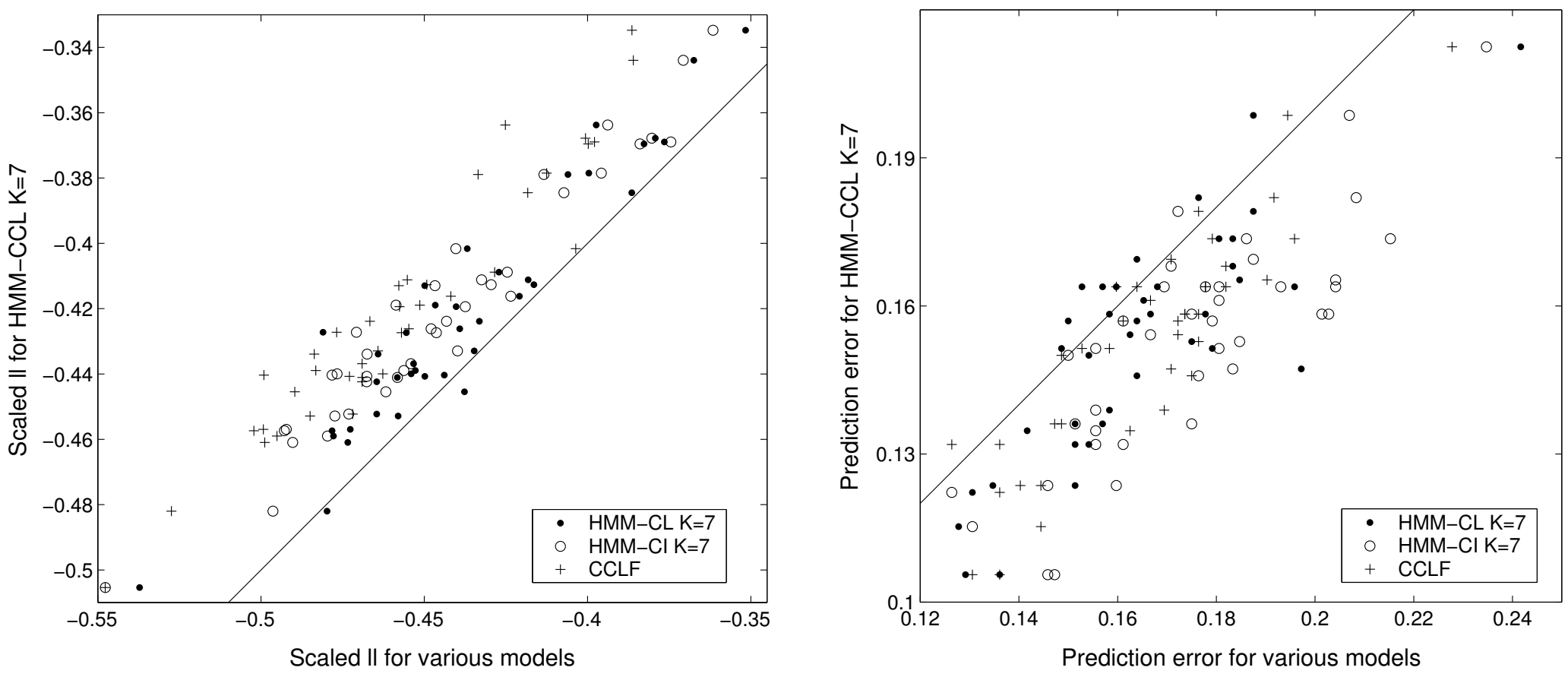

Figure 8: Western U.S. data: Scatterplots of out-of-sample scaled log-likelihoods (left) and average prediction error (right) obtained by leave-one-winter-out cross-validation. The line corresponds to $y = x$. The independent chains model is not shown since it is beyond the range of the plot (average $ll = -0.5204$, average error $= 0.221$).

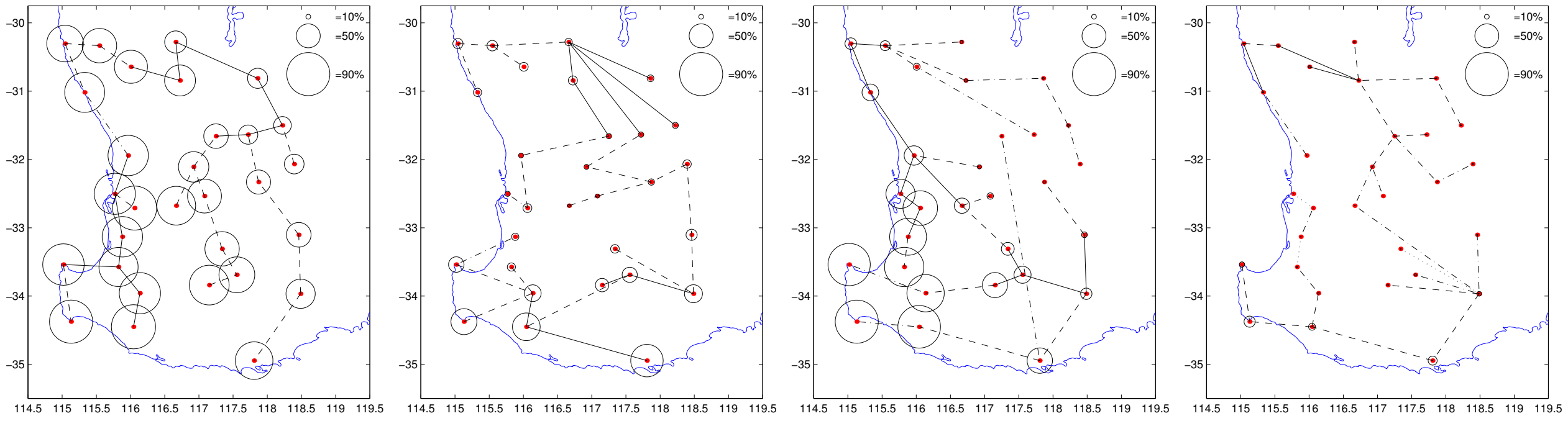

Figure 9: Graphical interpretation of the hidden states for a 5-state HMM-CL trained on Southwestern Australia data (4 out of 5 shown). Radii of the circles indicate the precipitation probability for each station given the state. Lines between the stations indicate the edges in the graph while different types of lines indicate the strength of mutual information of the edges.

Examples of the Chow-Liu tree structures learned by the model are shown in Figure 9 for the 5-state HMM-CL model trained on all 15 years of Southwestern Australia data. The states learned by the model correspond to a variety of wet and dry spatial patterns. The tree structures are consistent with the meteorology and topography of the region (Hughes et al. 1999). Winter rainfall over SW Australia is large-scale and frontal, impacting the southwest corner of the domain first and foremost. Hence, the tendency for correlations between stations along the coast during moderately wet weather states. Interesting correlation structures are also identified in the north of the domain even during dry conditions.

## 5 Conclusions

We have investigated a number of approaches for modeling multivariate discrete-valued time series. In particular we illustrated how Chow-Liu trees could be embedded within hidden Markov models to provide improved temporal and multivariate dependence modeling in a tractable and parsimonious manner. We also introduced the conditional Chow-Liu forest model, a natural extension of Chow-Liu trees for modeling conditional distributions such as multivariate data with temporal dependencies. Experimental results on real-world precipitation data indicate that these models provide systematic improvements over simpler alternatives such as assuming conditional independence of the multivariate outputs. There are a number of extensions that were not discussed in this paper but that can clearly be pursued, including (a) using informative priors over tree-structures (e.g., priors on edges based on distance and topography for precipitation station models), (b) models for real-valued or mixed data (e.g., modeling precipitation amounts as well as occurrences), (c) adding input variables to the HMMs (e.g., to model "forcing" effects from atmospheric measurements—for initial results see Robertson et al. (to appear)), and (d) performing systematic experiments comparing these models to more general classes of dynamic Bayesian networks where temporal and multivariate structure is learned directly.

**Acknowledgements**

We would like to thank Stephen Charles of CSIRO, Australia, for providing us with the Western Australia data. This work was supported in part by the Department of Energy under grant DE-FG02-02ER63413 and in part by the National Science Foundation under Grant No. SCI-0225642 as part of the OptIPuter project.